\documentclass[11pt]{article}

\usepackage{acl}

\usepackage{times}
\usepackage{latexsym}
\usepackage{graphicx}
\usepackage{multirow}
\usepackage{tabularx}
\usepackage{soul}
\usepackage{color}
\usepackage{xcolor}
\definecolor{darkgreen}{RGB}{83,129,53}
\definecolor{darkred}{RGB}{163,21,21}
\usepackage{amsmath}
\usepackage{float}
\usepackage{cleveref}
\usepackage[T1]{fontenc}

\usepackage[utf8]{inputenc}

\usepackage{microtype}

\title{Retrieval-Augmented Code Generation for Universal Information Extraction}

\author{Yucan Guo$^{1,2}$, Zixuan Li$^{1,2,}$\thanks{$^*$ Corresponding authors.} , Xiaolong Jin$^{1,2,*}$, Yantao Liu$^{1,2}$, Yutao Zeng$^{3}$, Wenxuan Liu$^{1,2}$,\\ {\bf Xiang Li$^{1,2}$, Pan Yang$^{3}$, Long Bai$^{1,2}$, Jiafeng Guo$^{1,2}$} \and {\bf Xueqi Cheng$^{1,2}$} \\
    \vspace{-0.1cm}
    {\normalsize $^1$ CAS Key Laboratory of Network Data Science and Technology,}\\
    \vspace{-0.1cm}
    {\normalsize Institute of Computing Technology, Chinese Academy of Sciences}\\
    \vspace{-0.1cm}
    {\normalsize $^2$ School of Computer Science and Technology, University of Chinese Academy of Sciences}\\
    \vspace{-0.2cm}
    {\normalsize $^3$ Platform and Content Group, Tencent}\\
    \vspace{-0.2cm}
    \texttt{\small \{guoyucan23z,lizixuan,jinxiaolong,liuyantao22s,liuwenxuan24z\}@ict.ac.cn}, \\
    \vspace{-0.2cm}
    \texttt{\small \{lixiang23s,bailong,guojiafeng,cxq\}@ict.ac.cn}, \\
    \vspace{-0.2cm}
    \texttt{\small tedzeng@tencent.com}, \texttt{\small im.panyang@gmail.com}
}

\begin{document}
\begin{sloppy}
\maketitle
\begin{abstract}

Information Extraction (IE) aims to extract structural knowledge (e.g., entities, relations, events) from natural language texts, which brings challenges to existing methods due to task-specific schemas and complex text expressions. Code, as a typical kind of formalized language, is capable of describing structural knowledge under various schemas in a universal way. On the other hand, Large Language Models (LLMs) trained on both codes and texts have demonstrated powerful capabilities of transforming texts into codes, which provides a feasible solution to IE tasks. Therefore, in this paper, we propose a universal retrieval-augmented code generation framework based on LLMs, called Code4UIE, for IE tasks. Specifically, Code4UIE adopts Python classes to define task-specific schemas of various structural knowledge in a universal way. By so doing, extracting knowledge under these schemas can be transformed into generating codes that instantiate the predefined Python classes with the information in texts. To generate these codes more precisely, Code4UIE adopts the in-context learning mechanism to instruct LLMs with examples. In order to obtain appropriate examples for different tasks, Code4UIE explores several example retrieval strategies, which can retrieve examples semantically similar to the given texts. Extensive experiments on five representative IE tasks across nine datasets demonstrate the effectiveness of the Code4UIE framework. 
\end{abstract}

\section{Introduction}
Information Extraction (IE) aims to transform natural language text into
structured knowledge, which is a critical task in Natural Language Processing
(NLP) that facilitates many downstream applications, such as knowledge graph construction, question-answering, and recommendation systems~\cite{Luan2018-multi,Yan2018-assertion,Guo2022-intelligent}.
IE presents two major challenges to existing models: various task-specific schemas and complex text expressions. Although all IE tasks (e.g., Named Entity Recognition (NER), Relation Extraction (RE)) are to extract structural knowledge, different tasks define different knowledge structures in terms of schemas, which cannot be readily presented in a universal manner.
Besides, due to the complexity and diversity of natural language, the same structural knowledge can be expressed in various ways by different synonyms, phrases, sentence structures, and writing styles. 
To tackle these two challenges, \citet{Lu2022-uie} designed Structural Schema Instructor (SSI) to describe the extraction target of a task, and Structured Extraction Language (SEL) to uniformly represent the extraction structure. They further proposed the first unified text-to-structure generation framework named UIE, for IE tasks, based on SSI and SEL. However, large-scale training is required for UIE, which leads to the poor performance of UIE in low-resource settings.

\begin{figure}[htbp]
    \centering
    \includegraphics[width=\linewidth]{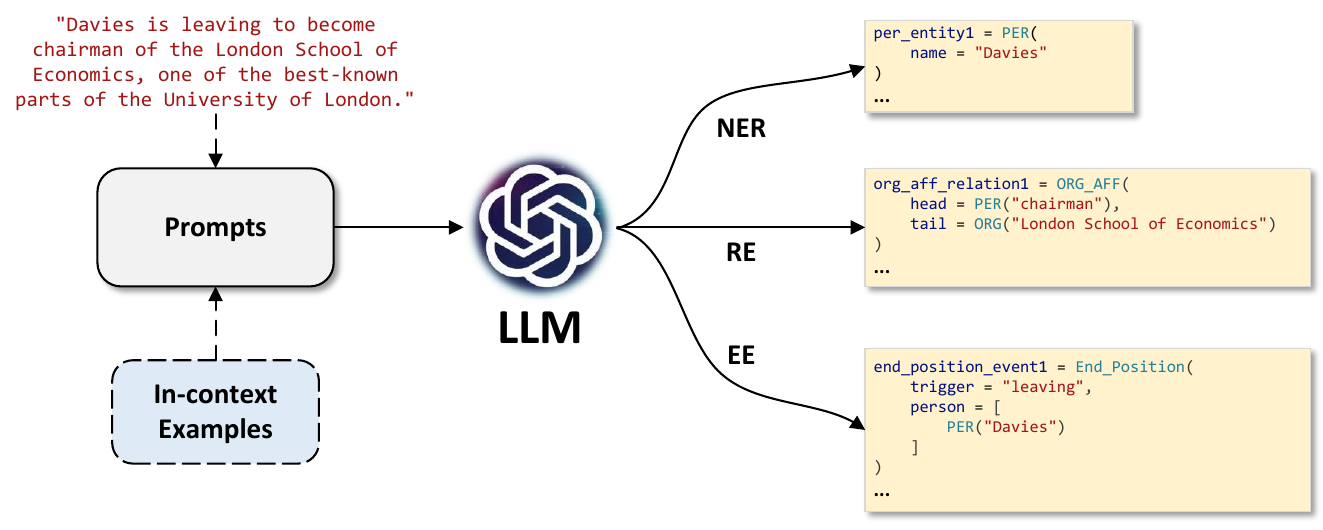}
    \caption{Illustration of transforming IE tasks into the code generation task.\label{fig:IE2CG}}
\end{figure}
Code, as a typical kind of formalized language, has the innate ability to precisely represent and define knowledge under various schemas, providing a universal and robust solution to the challenge of various schemas. Simultaneously, Large Language Models (LLMs) trained on vast amounts of text data expose the model to a rich diversity of language expressions and thus endow them with the capability to understand complex text expressions. Moreover, with both code and text as training data, LLMs have demonstrated powerful capabilities of transforming texts into code, which provides a possible solution to IE tasks. 
Motivated by these, in this paper, we propose a novel code generation framework to tackle these challenges for Universal Information Extraction, thus called Code4UIE. It leverages the power of LLMs to conduct code generation for IE tasks. The key idea behind Code4UIE is to use code features, such as inheritance, to represent the schemas required for IE. This provides a uniform mechanism for handling the diverse structures encountered in different IE tasks.

With the code-style schemas serving as its input, Code4UIE is capable of decoding heterogeneous IE structures as uniform code generation. To accomplish this, we design two types of code-style prompts that guide the model in generating appropriate codes. To understand complex text expressions, we enhance Code4UIE with unified example retrieval strategies, which help the model generalize better from the training examples and adapt to new tasks more effectively.

Extensive experiments conducted on five different IE tasks, including Named Entity Recognition (NER), Relation Extraction (RE), Event Detection (ED), Event Argument Extraction (EAE), and Event Extraction (EE), across nine datasets validate the effectiveness of Code4UIE.

In summary, our contributions are as follows:
\begin{enumerate}
    \item We propose a schema-based representation method that can universally define various task-specific schemas in IE tasks in terms of Python classes.  
    \item We transform IE tasks into a universal code generation task based on the schema representation method, and further propose a retrieval-augmented code generation framework based on LLMs, Code4UIE, to tackle this universal code generation task. Code4UIE is equipped with unified example retrieval strategies to help LLMs understand complex text expressions better by in-context learning.
    \item Experimental results on five IE tasks over nine datasets demonstrate that Code4UIE outperforms current LLMs-based IE approaches on all IE tasks.
\end{enumerate}

\section{Related Work}
\subsection{Universal Information Extraction (UIE)}
Information extraction is a critical task in natural language processing, aiming to extract structural information from unstructured or semi-structured texts. 
Traditional methods decompose IE into multiple sub-tasks and propose task-specific neural models accordingly, such as the ACE framework~\cite{Wang2021-automated} for NER task, PL-Marker~\cite{Ye2022-packed} for NER and RE tasks, DEGREE~\cite{Hsu2022-degree} for EAE and EE tasks.

To develop a universal IE architecture that can uniformly model different IE tasks, \citet{Lu2022-uie} proposed UIE, a framework that formulates four IE tasks as a unified text-to-structure generation task based on the structured extraction language they designed. Later, \citet{Lou2023-usm} proposed USM, an end-to-end unified semantic matching framework that decouples various IE tasks into token-token linking and token-label linking tasks. 
However, UIE requires separate fine-tuning for different downstream tasks, and USM has a long training and inference time. Very recently, to solve these two problems, \citet{Wang2023-instructuie} proposed InstructUIE, a UIE framework based on multi-task instruction tuning, reformulating IE tasks as a natural language generation problem.

Although a recent wave of UIE models has emerged, they typically require large-scale training to achieve performance comparable to task-specific models. In few-shot scenarios, they often exhibit subpar performance.

\subsection{LLM-based Information Extraction}
With the rise of large language models~\cite{Brown2020-language,ouyang2022-training,Touvron2023-llama} and their demonstrated strong few-shot learning capabilities, some recent research has focused on few-shot information extraction based on LLMs. \citet{Li2023-Evaluating} conducted systematical experiments with text-style prompts to evaluate the overall IE capabilities of ChatGPT and found that ChatGPT’s performance in the Standard-IE setting is poor. \citet{dyer-2023-revisiting} explored the performance of LLMs on the RE task also with a text-style prompt and found that few-shot prompting with GPT-3 achieves near SOTA performance after manually re-evaluating the output of LLMs. 

More recently, there has been a trend of using LLMs to transform information extraction tasks into code generation tasks.\citet{Li2023-codeie} proposed CodeIE, an approach that recasts the NER and RE tasks into code completion tasks for Python functions with Code-LLMs. \citet{Wang2023-code4struct} proposed a method named Code4Struct to convert the EAE task into a Python code generation task and rivaled the SOTA on 20-shot data.

While there have been various attempts to apply LLMs to different IE tasks, there is still a gap in utilizing LLMs for UIE as of now.

\section{Code4UIE}
Our proposed framework, Code4UIE, as shown in \Cref{fig:model}, adopts Python classes to represent task-specific schemas and the in-context learning mechanism to instruct LLMs with retrieved examples. In this section, we first introduce the schema representation method in \Cref{sec:schema representation}. Then, the structure of the input prompt and retrieval strategies designed for retrieving in-context examples are described in \Cref{sec:prompt construction} and \Cref{sec:retrieval strategies}, respectively. 

\subsection{Schema Representation}
\label{sec:schema representation}
To represent the schema of different datasets and various information extraction tasks in a unified way, we design a series of inheritable Python classes systematically, including a base class and a set of hierarchical subclasses for each type of structured information (i.e., entity, relation, and event). 

\subsubsection{Entity Definition}
To define entities by Python classes, a Python base class, "Entity", is served as the parent class for all entity classes, with each specific entity type corresponding to a subclass of "Entity". 
\begin{figure}[htbp]
    \centering
    \includegraphics[width=\linewidth]{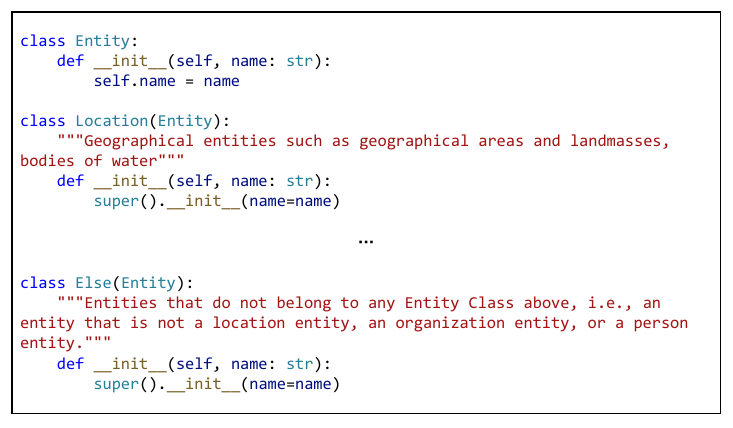}
    \caption{An example of the base class and subclass definitions of entities.\label{fig:entity class definitions}}
\end{figure}
\subsubsection{Relation Definition}
In the relation definition part, the abstract class "Relation" is inherited by other specific relation subclasses. As shown in \Cref{fig:relation class definitions}, each relation subclass has two variables, namely {\tt self.head} and {\tt self.tail}, which represent a relation triplet {\tt (self.head, Name\_of\_the\_Subclass, self.tail)}.
\begin{figure}[htbp]
    \centering
    \includegraphics[width=\linewidth]{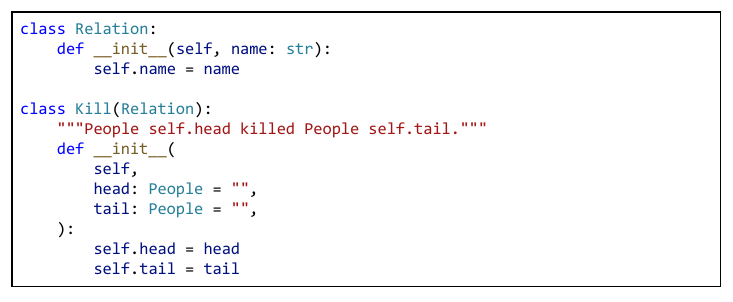}
    \caption{An example of the base class and subclass definitions of relations.\label{fig:relation class definitions}}
\end{figure}

\subsubsection{Event Definition}
The event definition part consists of the definition of the base class "Event", the definition of event trigger, and definitions of event subclasses. As shown in \Cref{fig:event class definitions}, trigger class and event subclasses are illustrated by natural language appearing in the form of annotation. Each event subclass has two kinds of variables, i.e., trigger and event arguments, in which the trigger is a string, and arguments with the same argument role (e.g., {\tt purpose} of {\tt Databreach}) are represented by a list of strings.
\begin{figure}[htbp]
    \centering
    \includegraphics[width=\linewidth]{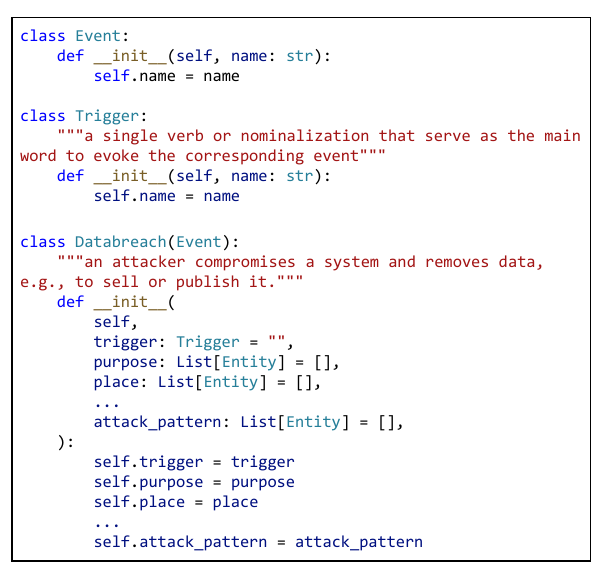}
    \caption{An example of the base class and subclass definitions of events.\label{fig:event class definitions}}
\end{figure}
\begin{figure*}[htbp]
  \centering
  \includegraphics[width=\textwidth]{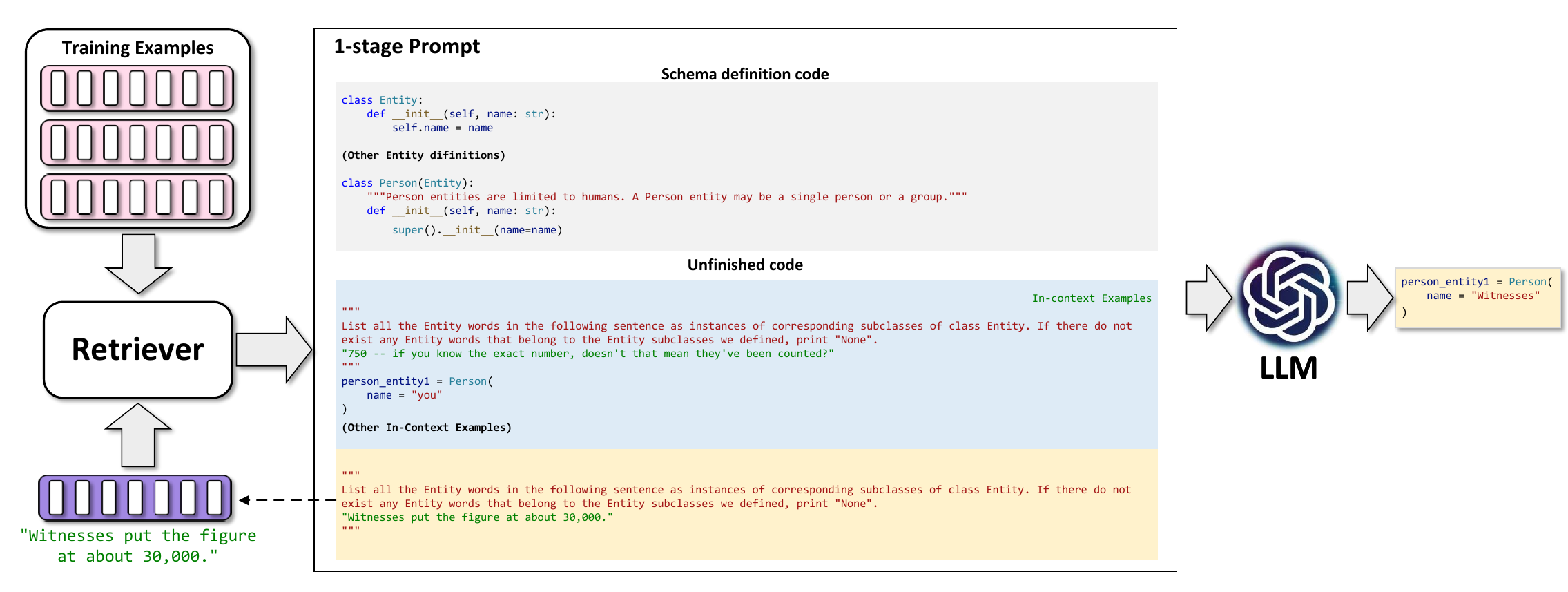}
  \caption{An illustration diagram of the proposed Code4UIE framework.}
  \label{fig:model}
\end{figure*}

\subsection{Prompt Construction}
\label{sec:prompt construction}
Based on the schema representation method we introduced in \Cref{sec:schema representation}, each information mention (e.g., entity mention, relation mention, and event mention) can be represented by an instance of a specific Python class. By this means, all the schema and structural information can be represented in a scalable way, and all the IE tasks can be transformed into a unified code generation task, i.e., generating instances of a series of predefined Python classes. To enable large language models to effectively complete this code generation task, we designed a 1-stage prompt and a 2-stage prompt.

\subsection{1-stage Prompt}
Code4UIE will extract the information in an end-to-end manner with a 1-stage code-style prompt, which consists of schema definition code and unfinished code. The schema definition code should contain all class definitions related to the task and dataset. For the NER task, entity definitions are required; for the RE task, entity definitions and relation definitions are required; for the ED, EAE, and EE tasks, entity definitions and event definitions are required. The beginning of the unfinished code is $k$ in-context examples, each of them consists of a task instruction in the form of an annotation and the correct output code in the form of instances of Python classes in the schema definition code. Following this is a task instruction about the input sentence that leaves the LLM to generate the output code. The structure of the task instruction annotation is shown in \Cref{fig:1-stage task instruction}, and an example of the 1-stage prompt can be found in \Cref{fig:model}.
\begin{figure}[h]
    \centering
    \includegraphics[width=\linewidth]{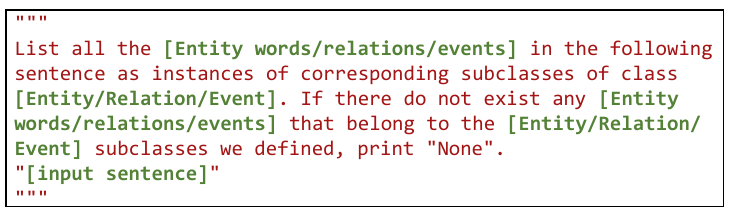}
    \caption{The structure of the task instruction annotation in the 1-stage prompt.\label{fig:1-stage task instruction}}
\end{figure}

\subsection{2-stage Prompt}
With a 2-stage prompt, Code4UIE will identify types of entities/relations/events in Stage 1, and extract the exact entities/relations/events in Stage 2 based on the output of Stage 1. Note that the ED and EAE tasks do not have a 2-stage prompt as ED only needs to identify event types, and EAE only needs to extract event arguments of a given event type.
\begin{figure*}
    \centering
    \includegraphics[width=\linewidth]{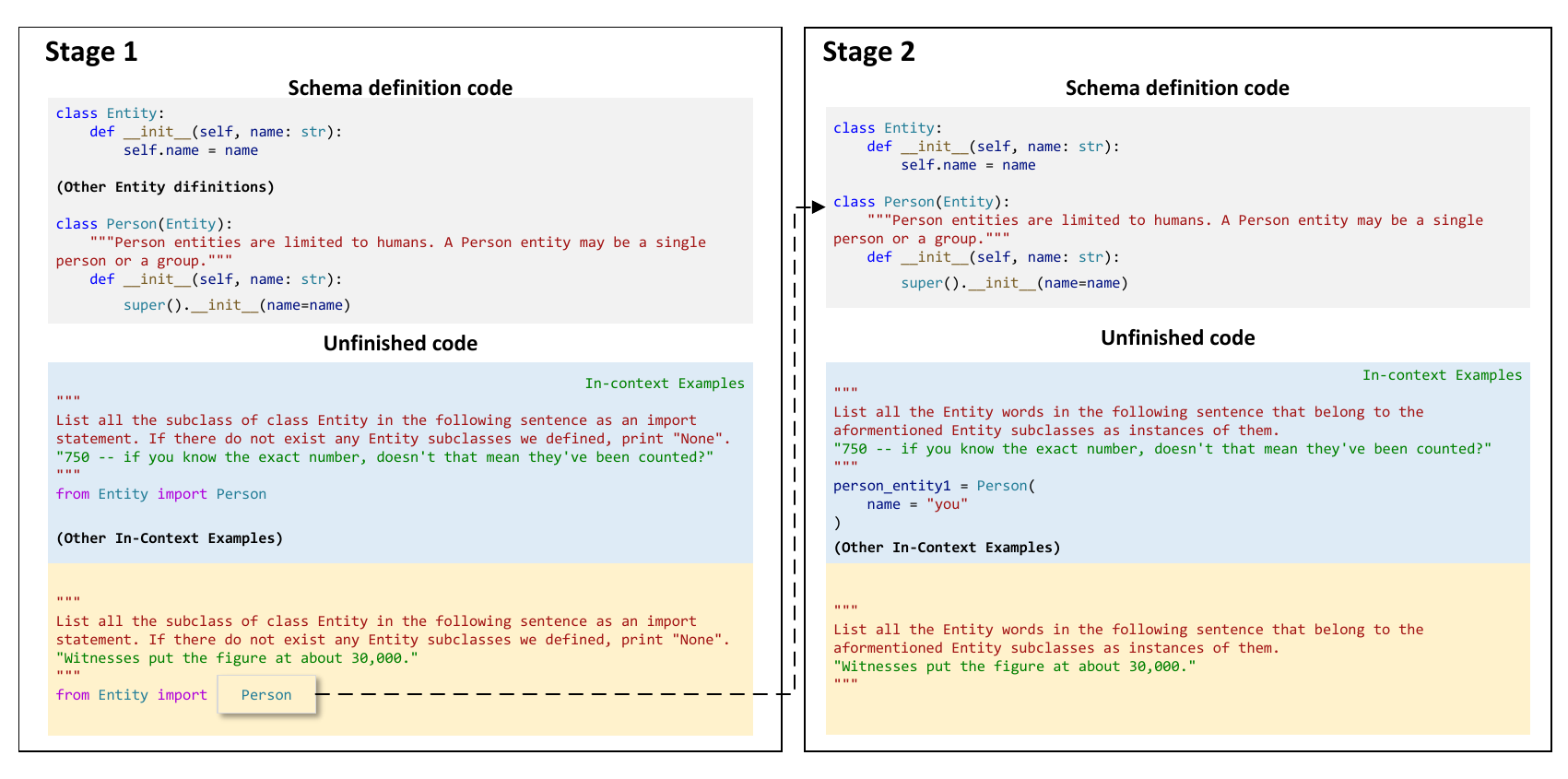}
    \caption{An illustration of the 2-stage prompt.}
    \label{fig:2-stage prompt}
\end{figure*}
\Cref{fig:2-stage prompt} shows an example of the 2-stage prompt, though the prompt in each stage still consists of schema definition code and unfinished code, there are two main differences between 1-stage prompt and 2-stage prompt: 
\begin{enumerate}
    \item[(1)] The code generation task in Stage 1 is to complete an import statement (e.g., {\tt from Entity import \textcolor{darkgreen}{\textbf{Person}}}) instead of generating instances of Python class.
    \item[(2)] The schema definition code in Stage 2 only contains those classes related to the import statement completed in Stage 1 (e.g., {\tt class \textcolor{darkgreen}{\textbf{Entity}}} and {\tt class \textcolor{darkgreen}{\textbf{Person}}}), while that of the 1-stage prompt and Stage 1 prompt of the 2-stage prompt should contain all the Python classes related to the task-specific schema.
\end{enumerate}

\subsection{In-context Example Retrieval Strategies}
\label{sec:retrieval strategies}
The selection of in-context examples plays a critical role in the few-shot in-context learning scenario. To find examples that suit the Code4UIE prompt structure, the specific IE task, and the input sentence better, Code4UIE proposes two kinds of example retrieval strategies, namely sentence embedding-based retrieval strategy and anonymous sentence embedding-based retrieval strategy.
\subsubsection{Sentence Embedding-based Retrieval Strategy}
\label{sec:sent-embed retrieval strategy}
The sentence embedding-based retrieval strategy uses MPNet~\cite{Song2020-MPNet} to encode the input sentence and sentences in the training set. The $k$ samples with the highest similarity to the input sentence in the training set will be selected as in-context examples, in which the similarity calculation is based on Euclidean distance.
This strategy is employed and tested in all five IE tasks. 

\subsubsection{Anonymous Sentence Embedding-based Retrieval Strategy}
\label{sec:anony-sent-smbed retrieval strategy}
Considering that selecting samples with higher similarity in terms of context and entity types, rather than specific entity text information, is more effective for complex IE tasks like EE, we propose an anonymous sentence embedding-based retrieval strategy. The key idea of this anonymous strategy is to use a NER model, FLERT~\cite{Schweter2020-FLERT}, to replace entities in a sentence by their entity types when encoding, and utilizing that anonymous information for the calculation of sentence similarity when performing in-context example retrieval.
This strategy is employed and tested in all IE tasks except the NER task as an existing NER model is used during the retrieval process. 

\section{Experiments}
\begin{table*}[htbp]
    \centering
    \renewcommand{\arraystretch}{1.1}
    \resizebox{\linewidth}{!}
    {
    \begin{tabular}{c c c c c c c}
    \hline
        \multirow{2}{*}{\bf Model} & \multirow{2}{3cm}{\centering \bf Prompt \\ Type} & \multirow{2}{3cm}{\centering \bf Retrieval \\ Strategy} & \multirow{2}{*}{\bf ACE04} & \multirow{2}{*}{\bf ACE05} & \multirow{2}{*}{\bf CoNLL03} \\
        & & & & & \\
    \hline
        UIE-large~\cite{Lu2022} & text & - & {\bf 86.89} & 85.78 & {\bf 92.99} \\
        InstructUIE~\cite{Wang2023} & text & - & - & {\bf 86.66} & 92.94 \\
    \hline
        \multirow{2}{6cm}{\centering code-davinci-002 (CodeIE)\\ \cite{Li2023-codeie}} & text & semi-random & 49.58 (16) & 49.55 (16) & 72.66 (25) \\
         & code & semi-random & 55.29 (16) & 54.82 (16) & 82.32 (25) \\
    \hline 
        \multirow{2}{*}{gpt-3.5-turbo-16k} & 1-stage code & semi-random & 47.6 (16) & 48.2 (16) & 73.0 (25) \\
         & 2-stage code & semi-random & 45.9 (16) & 48.4 (16) & 72.1 (25) \\
    \hline
        gpt-3.5-turbo-16k (Code4UIE) & 1-stage code & sent-embed & 54.0 (16) & 57.0 (16) & 79.7 (25) \\
    \hline
        text-davinci-002 (Code4UIE) & 1-stage code & sent-embed & 55.8 (10) & 58.2 (10) & 82.1 (20)\\
    \hline
        text-davinci-003 (Code4UIE) & 1-stage code & sent-embed & {\bf 60.1} (10) & {\bf 60.9} (10) & {\bf 83.6} (20)\\
    \hline
    \end{tabular}
}
    \caption{Experimental results on the NER task, where retrieval strategy indicates the strategy used for retrieving in-context examples, and (\#) after the F1-score indicates the number of in-context examples.}
    \label{tab:NER performance}
\end{table*}
\subsection{Setup}
{\bf LLMs}\hspace{0.5em}
We adopt {\tt text-davinci-002}, {\tt text-davinci-003}, and {\tt gpt-3.5-turbo\\-16k} from OpenAI as the LLMs used in the experiments. The NER, RE and EAE tasks are mainly conducted with the former two LLMs that support 4k input tokens at most. The ED, EE, and some over-length tasks are conducted with the last LLM that supports at most 16k input tokens.
We access these LLMs through OpenAI API~\footnote{https://openai.com/product} and set the temperature $t=0$.

\noindent{\bf Datasets}\hspace{0.5em}
For the NER task, we evaluate the performance of models on ACE04~\cite{Doddington04-ace04}, ACE05~\cite{Walker-ace2005}, and CoNLL03~\cite{Sang2003-conll03}; for the RE task, we evaluate that on ADE~\cite{Gurulingappa2012-ADE}, ACE05~\cite{Walker-ace2005}, CoNLL04~\cite{Roth2004-conll04}, and NYT~\cite{Riedel2010-NYT}; for the ED, EAE and EE tasks, we evaluate that on ACE05~\cite{Walker-ace2005} and CASIE~\cite{Satyapanich2020-CASIE}.

\noindent{\bf Baselines}\hspace{0.5em}
We compare Code4UIE with two fully supervised universal information extraction models, a fully supervised event argument extraction model, a text generation-based relation extraction method, and two code generation-based information extraction methods:
\begin{itemize}
    \item {\bf UIE-large}~\cite{Lu2022-uie} is a universal information extraction model that formulates IE tasks as text-to-structure problems.
    \item {\bf InstructUIE}~\cite{Wang2023-instructuie} is a unified information extraction model based on multi-task instruction tuning.
    \item {\bf DEGREE}~\cite{Hsu2022-degree} is the current SOTA model on the EAE task that formulates event argument extraction as a conditional generation problem.
    \item {\bf In-context GPT3}~\cite{Wadhwa2023-revisiting} is a method to treat RE as a conditional text generation task.
    \item {\bf CodeIE}~\cite{Li2023-codeie} is a method to convert NER and RE tasks into code generation tasks and use Code-LLMs to solve these tasks.
    \item {\bf Code4Struct}~\cite{Wang2023-code4struct} is a method that formulates the EAE task as a code generation task for LLMs.
\end{itemize}
\begin{table*}[htbp]
    \centering
    \renewcommand{\arraystretch}{1.1}
    \resizebox{1\linewidth}{!}
    {
    \begin{tabular}{c c c c c c c}
    \hline
        \multirow{2}{*}{\bf Model} & \multirow{2}{3cm}{\centering \bf Prompt \\ Type} & \multirow{2}{3cm}{\centering \bf Retrieval \\ Strategy} & \multirow{2}{*}{\bf ADE} & \multirow{2}{*}{\bf ACE05} & \multirow{2}{*}{\bf CoNLL04} & \multirow{2}{*}{\bf NYT}\\
        & & & & & &\\
    \hline
        UIE-large~\cite{Lu2022} & text & - & - & {\bf 66.06} & 75.00 & - \\
        InstructUIE~\cite{Wang2023} & text & - & {\bf 82.31} & - & {\bf 78.48} & {\bf 90.47}\\
    \hline
        text-davinci-002 (In-context GPT3) & \multirow{2}{*}{text} & \multirow{2}{*}{fixed} & \multirow{2}{*}{51.5 (12)} & \multirow{2}{*}{-} & \multirow{2}{*}{31.8 (12)} & \multirow{2}{*}{17.6 (20)}\\
        \cite{Wadhwa2023-revisiting} & & & & & & \\
    \hline 
        \multirow{2}{5.5cm}{\centering code-davinci-002 (CodeIE)\\ \cite{Li2023-codeie}} & text & semi-random & - & 10.08 (14) & 47.30 (25) & 24.63 (24)\\
         & code & semi-random & - & 14.02 (14) & 53.10 (25) & 32.17 (24)\\
    \hline 
        \multirow{2}{*}{text-davinci-002} & 1 | 2-stage code & fixed & 56.5 (12) | - & - | - & 38.2 | 37.1 (12) & - | 21.0 (20)\\
        & 1 | 2-stage code & semi-random & - | - & 7.9 | 4.1 (14) & 47.9 | 45.9 (25) & overlength (24)\\
        \hline
        \multirow{2}{5.5cm}{\centering text-davinci-002 (Code4UIE)} & 1 | 2-stage code & sent-embed & 60.0 (12) | - & {\bf 18.6} (14) | - & 54.3 (12) | - & - | {\bf 62.1} (15)\\
        & 1 | 2-stage code & anony-sent-embed & 58.6 (12) | - & 17.5 (14) | - & 54.4 (12) | - & - | 54.4 (15)\\
    \hline
        \multirow{2}{*}{\centering gpt-3.5-turbo-16k (Code4UIE)} & 1-stage code & sent-embed & 65.3 (12) & 11.5 (14) & {\bf 54.6} (12) & 51.7 (20)\\
         & 1-stage code & anony-sent-embed & {\bf 63.6} (12) & 14.1 (14) & 44.3 (12) & 46.5 (20)\\
    \hline
    \end{tabular}
    }
    \caption{Experimental results on the RE task, where retrieval strategy indicates the strategy used for retrieving in-context examples, and (\#) after the F1-score indicates the number of in-context examples.}
    \label{tab:RE performance}
\end{table*}
\begin{table*}[htbp]
    \centering
    \renewcommand{\arraystretch}{1.1}
    \resizebox{1\linewidth}{!}
    {
    \begin{tabular}{c c c c c}
    \hline
        \multirow{2}{*}{\bf Model} & \multirow{2}{4cm}{\centering \bf Prompt \\ Type} & \multirow{2}{3cm}{\centering \bf Retrieval \\ Strategy} & \multirow{2}{*}{\bf ACE05} & \multirow{2}{*}{\bf CASIE}\\
        & & & &\\
    \hline
        UIE-large~\cite{Lu2022} & text & - & 73.36 & {\bf 69.33}\\
        InstructUIE~\cite{Wang2023} & text & - & {\bf 77.13} & 67.80\\
    \hline
        gpt-3.5-turbo-16k & text (follow the InstructUIE format) & random & 10.5 (10) & 26.6 (10)\\
    \hline 
        gpt-3.5-turbo-16k & 1-stage code & random & 15.5 (10) & 15.3 (10)\\
    \hline
        \multirow{2}{4cm}{\centering gpt-3.5-turbo-16k\\ (Code4UIE)} & 1-stage code & sent-embed & 35.4 (10) & 27.8 (10)\\
         & 1-stage code & anony-sent-embed & {\bf 37.4} (10) & {\bf 28.7} (10)\\
    \hline
    \end{tabular}
    }
    \caption{Experimental results on the ED task, where retrieval strategy indicates the strategy used for retrieving in-context examples, and (\#) after the F1-score indicates the number of in-context examples.}
    \label{tab:ED performance}
\end{table*}

\noindent{\bf Evaluation metrics}\hspace{0.5em}
We use {\bf Entity F1-score} as the evaluation metric for the NER task, {\bf Relation Strict F1-score} for the RE task, {\bf Trigger F1-score} for the ED task, and {\bf Argument F1-score} for the EE task following previous works~\cite{Li2023-codeie, Li2023-Evaluating}. Under these metrics, an entity/relation/event prediction is correct if its type and offsets are correct. As for the EAE task, {\bf Argument Head F1-score} is used as the evaluation metric following prior work~\cite{Wang2023-code4struct}. 
Under the {\bf Argument Head F1-score} evaluation metric, an argument is correctly identified when the head word span of predicted text matches that of the human-annotated text, rather than the requirement of complete text consistency of the {\bf Argument F1-score}.

\subsection{NER Performance}
\label{sec:NER performance}
We mainly compare the NER performance of Code4UIE with that of CodeIE, the current state-of-the-art LLM-based NER method, in this section. CodeIE uses {\tt code-davinci-002}, a GPT-3.5 model that has been optimized for code-completion tasks and supports at most 8k input tokens, in their experiment. However, this model is now deprecated by OpenAI, and we choose some GPT-3.5 text models that are weaker in code-completion tasks to ensure fairness. The performance of Code4UIE on the NER task is shown in \Cref{tab:NER performance}, from which we make the following observations:
\begin{enumerate}
    \item[(1)] Although Code4UIE performs worse than those fully supervised UIE models, it achieves better results with fewer in-context examples in comparison with CodeIE when using {\tt text-davinci-002} and {\tt text-davinci-003}.
    \item[(2)] The semi-random retrieval strategy (i.e., randomly select $k$ examples from each entity/relation type) that CodeIE adopted is not suitable for Code4UIE. This is reasonable because CodeIE does not define the schema in the prompt. Therefore, it relies on in-context examples of each type to assist the LLM in understanding the task, while Code4UIE has a clearly defined schema in the prompt, making it less suitable for the method of randomly selecting examples from each type.
    \item[(3)] Compared to the semi-random strategy, the sentence embedding-based retrieval strategy has been shown to improve the performance of Code4UIE by approximately 7\%, which demonstrates the effectiveness of the retrieval strategy that Code4UIE adopted.
\end{enumerate}

\subsection{RE Performance}
For the RE task, we mainly compare Code4UIE with the text generation-based RE method In-context GPT3, and the code generation-based method CodeIE. 
The metrics reported by In-context GPT-3 have undergone manual re-evaluating and are higher than the relation strict F1-score. To ensure fairness, we have reproduced their experimental results and calculated the strict matching metrics accordingly.
\Cref{tab:RE performance} shows the experimental result of different methods on the RE task and we summarize our findings as follows:
\begin{enumerate}
    \item [(1)] Our code-style prompts surpass In-context GPT3 with the same fixed examples that In-context GPT3 reported, proving the superiority of code-style prompts over text-style prompts. This advantage is further amplified by Code4UIE, which is equipped with two retrieval strategies we proposed.
    \item [(2)] Code4UIE performs better than CodeIE with fewer in-context examples, even on less code-savvy LLMs (i.e., {\tt text-davinci-002} and {\tt gpt-3.5-turbo-16k}). However, there is still a performance gap when comparing Code4UIE with those fully supervised UIE models.
    \item [(3)] The sentence embedding-based and anonymous sentence embedding-based retrieval strategies are more suitable for Code4UIE, which is consistent with results in \Cref{sec:NER performance}. It also seems that the sentence embedding-based retrieval strategy is a better choice for the RE task.
\end{enumerate}
\begin{table*}[htbp]
    \centering
    \renewcommand{\arraystretch}{1.1}
    \resizebox{.8\linewidth}{!}
    {
    \begin{tabular}{c c c c}
    \hline
        \multirow{2}{*}{\bf Model} & \multirow{2}{3cm}{\centering \bf Prompt \\ Type} & \multirow{2}{3cm}{\centering \bf Retrieval \\ Strategy} & \multirow{2}{*}{\bf ACE05} \\
        & & &\\
    \hline
        DEGREE~\cite{Hsu2022-degree} & text & - & {\bf 73.5}\\
    \hline
        \multirow{3}{4cm}{\centering text-davinci-002\\ (Code4Struct)\\ \cite{Wang2023-code4struct}} & text & random & 48.9 (10)\\
         & code & random & 53.9 (10)\\
         & code & sibling & 54.9 (10)\\
    \hline 
        text-davinci-002 & 1-stage code & sibling & 54.3 (10)\\
    \hline
        \multirow{2}{4cm}{\centering text-davinci-002\\ (Code4UIE)} & 1-stage code & sent-embed & 56.7 (10)\\
         & 1-stage code & anony-sent-embed & {\bf 57} (10)\\
    \hline
    \end{tabular}
    }
    \caption{Experimental results on the EAE task, where retrieval strategy indicates the strategy used for retrieving in-context examples, and (\#) after the F1-score indicates the number of in-context examples.}
    \label{tab:EAE performance}
\end{table*}
\begin{table*}[htbp]
    \centering
    \renewcommand{\arraystretch}{1.1}
    \resizebox{1\linewidth}{!}
    {
    \begin{tabular}{c c c c c}
    \hline
        \multirow{2}{*}{\bf Model} & \multirow{2}{4cm}{\centering \bf Prompt \\ Type} & \multirow{2}{3cm}{\centering \bf Retrieval \\ Strategy} & \multirow{2}{*}{\bf ACE05} & \multirow{2}{*}{\bf CASIE}\\
        & & & &\\
    \hline
        UIE-large~\cite{Lu2022} & text & - & 54.79 & 61.30\\
        InstructUIE~\cite{Wang2023} & text & - & {\bf 72.94} & {\bf 63.53}\\
    \hline
        gpt-3.5-turbo-16k & text (follow the InstructUIE format) & random & 4.5 (10) & 11.3 (10)\\
    \hline 
        gpt-3.5-turbo-16k & 2-stage code & random & 11.0 (10) & 15.3 (10)\\
    \hline
        \multirow{2}{4cm}{\centering gpt-3.5-turbo-16k\\(Code4UIE)} & 2-stage code & sent-embed & 21.2 (10) & 29.5 (10)\\
         & 2-stage code & anony-sent-embed & {\bf 21.3} (10) & {\bf 30.8} (10)\\
    \hline
    \end{tabular}
    }
    \caption{Experimental results on the EE task, where retrieval strategy indicates the strategy used for retrieving in-context examples, and (\#) after the F1-score indicates the number of in-context examples.}
    \label{tab:EE performance}
\end{table*}
\subsection{ED Performance}
\label{sec:ED performance}
To compare our code-style prompt with the text-style prompt, we designed a text prompt following the instruction format of InstructUIE~\cite{Wang2023-instructuie}. Experiment results in \Cref{tab:ED performance} show that the performance of the code-style prompt surpasses that of the text-style prompt, and Code4UIE outperforms the text prompt even more significantly. Specifically, Code4UIE yields improvements of 24.9\% and 26.9\% on the ACE05 dataset with the sentence embedding-based retrieval strategy and the anonymous sentence embedding-based retrieval strategy, respectively.
For the code-style prompt itself, the F1-score increases marginally by around 20\% when replacing randomly chosen examples with examples that our retrieval strategies found, which again proves the effectiveness of in-context example retrieval strategies that Code4UIE adopted. 

\subsection{EAE Performance}
Results of EAE are shown in \Cref{tab:EAE performance}, with the same {\tt text-davinci-002} model and an equal number of in-context examples, the performance of Code4UIE using sentence embedding-based retrieval strategy and anonymous sentence embedding-based retrieval strategy surpasses Code4Struct in both cases.

\subsection{EE Performance}
Similar to \Cref{sec:ED performance}, we construct a text-style prompt for the EE task for comparison. \Cref{tab:EE performance} shows the performance of text-style prompt and code-style prompt on the EE task. Code4UIE outperforms the text prompt by a large margin (about 16\% - 19\% gain in the F1-score) with both sentence embedding-based retrieval strategy and anonymous sentence embedding-based retrieval strategy. From \Cref{tab:ED performance}-\Cref{tab:EE performance}, we can also see that the anonymous sentence embedding-based retrieval strategy is a better choice for event-related tasks rather than the sentence embedding-based retrieval strategy, demonstrating that context and entity type information play a more important role in the process of in-context learning for complicated IE tasks such as ED, EAE and EE.

\section{Conclusion}
In this paper, we proposed a retrieval-augmented code generation framework for universal information extraction, namely Code4UIE, which formulates different IE tasks as a unified Python instance code completion task. Code4UIE adopts two effective in-context example retrieval strategies, which enable it to retrieve in-context examples that are better suited for the current sentence and task, resulting in a significant improvement in IE performance. Experimental results demonstrate that Code4UIE surpasses current LLMs-based IE approaches on all IE tasks.


\bibliography{anthology,custom}

\begin{thebibliography}{28}
\expandafter\ifx\csname natexlab\endcsname\relax\def\natexlab#1{#1}\fi

\bibitem[{Brown et~al.(2020)Brown, Mann, Ryder, Subbiah, Kaplan, Dhariwal,
  Neelakantan, Shyam, Sastry, Askell, Agarwal, Herbert-Voss, Krueger, Henighan,
  Child, Ramesh, Ziegler, Wu, Winter, Hesse, Chen, Sigler, Litwin, Gray, Chess,
  Clark, Berner, McCandlish, Radford, Sutskever, and
  Amodei}]{Brown2020-language}
Tom Brown, Benjamin Mann, Nick Ryder, Melanie Subbiah, Jared~D Kaplan, Prafulla
  Dhariwal, Arvind Neelakantan, Pranav Shyam, Girish Sastry, Amanda Askell,
  Sandhini Agarwal, Ariel Herbert-Voss, Gretchen Krueger, Tom Henighan, Rewon
  Child, Aditya Ramesh, Daniel Ziegler, Jeffrey Wu, Clemens Winter, Chris
  Hesse, Mark Chen, Eric Sigler, Mateusz Litwin, Scott Gray, Benjamin Chess,
  Jack Clark, Christopher Berner, Sam McCandlish, Alec Radford, Ilya Sutskever,
  and Dario Amodei. 2020.
\newblock \href
  {https://proceedings.neurips.cc/paper_files/paper/2020/file/1457c0d6bfcb4967418bfb8ac142f64a-Paper.pdf}
  {Language models are few-shot learners}.
\newblock In \emph{Advances in Neural Information Processing Systems},
  volume~33, pages 1877--1901. Curran Associates, Inc.

\bibitem[{Christopher et~al.(2006)Christopher, Stephanie, Julie, and
  Kazuaki}]{Walker-ace2005}
Walker Christopher, Strassel Stephanie, Medero Julie, and Maeda Kazuaki. 2006.
\newblock Ace 2005 multilingual training corpus ldc2006t06.
\newblock In \emph{Philadelphia: Linguistic Data Consortium}. Web Download.

\bibitem[{Doddington et~al.(2004)Doddington, Mitchell, Przybocki, Ramshaw,
  Strassel, and Weischedel}]{Doddington04-ace04}
George~R. Doddington, Alexis Mitchell, Mark~A. Przybocki, Lance~A. Ramshaw,
  Stephanie~M. Strassel, and Ralph~M. Weischedel. 2004.
\newblock \href {http://www.lrec-conf.org/proceedings/lrec2004/summaries/5.htm}
  {The automatic content extraction {(ACE)} program - tasks, data, and
  evaluation}.
\newblock In \emph{Proceedings of the Fourth International Conference on
  Language Resources and Evaluation, {LREC} 2004, May 26-28, 2004, Lisbon,
  Portugal}. European Language Resources Association.

\bibitem[{Dyer(2023)}]{dyer-2023-revisiting}
Andrew~Thomas Dyer. 2023.
\newblock \href {https://doi.org/10.18653/v1/2023.sigtyp-1.11} {Revisiting
  dependency length and intervener complexity minimisation on a parallel corpus
  in 35 languages}.
\newblock In \emph{Proceedings of the 5th Workshop on Research in Computational
  Linguistic Typology and Multilingual NLP}, pages 110--119, Dubrovnik,
  Croatia. Association for Computational Linguistics.

\bibitem[{Guo et~al.(2022)Guo, Wang, Zhao, Diao, Chen, Ding, He, Lu, Xiao,
  Long, Yu, and Wu}]{Guo2022-intelligent}
Xiaojie Guo, Shugen Wang, Hanqing Zhao, Shiliang Diao, Jiajia Chen, Zhuoye
  Ding, Zhen He, Jianchao Lu, Yun Xiao, Bo~Long, Han Yu, and Lingfei Wu. 2022.
\newblock \href {https://doi.org/10.1609/aaai.v36i11.21501} {Intelligent online
  selling point extraction for e-commerce recommendation}.
\newblock \emph{Proceedings of the AAAI Conference on Artificial Intelligence},
  36(11):12360--12368.

\bibitem[{Gurulingappa et~al.(2012)Gurulingappa, Rajput, Roberts, Fluck,
  Hofmann-Apitius, and Toldo}]{Gurulingappa2012-ADE}
Harsha Gurulingappa, Abdul~Mateen Rajput, Angus Roberts, Juliane Fluck, Martin
  Hofmann-Apitius, and Luca Toldo. 2012.
\newblock \href {https://doi.org/https://doi.org/10.1016/j.jbi.2012.04.008}
  {Development of a benchmark corpus to support the automatic extraction of
  drug-related adverse effects from medical case reports}.
\newblock \emph{Journal of Biomedical Informatics}, 45(5):885--892.
\newblock Text Mining and Natural Language Processing in Pharmacogenomics.

\bibitem[{Hsu et~al.(2022)Hsu, Huang, Boschee, Miller, Natarajan, Chang, and
  Peng}]{Hsu2022-degree}
I-Hung Hsu, Kuan-Hao Huang, Elizabeth Boschee, Scott Miller, Prem Natarajan,
  Kai-Wei Chang, and Nanyun Peng. 2022.
\newblock \href {https://doi.org/10.18653/v1/2022.naacl-main.138} {{DEGREE}: A
  data-efficient generation-based event extraction model}.
\newblock In \emph{Proceedings of the 2022 Conference of the North American
  Chapter of the Association for Computational Linguistics: Human Language
  Technologies}, pages 1890--1908, Seattle, United States. Association for
  Computational Linguistics.

\bibitem[{Li et~al.(2023{\natexlab{a}})Li, Fang, Yang, Wang, Ye, Zhao, and
  Zhang}]{Li2023-Evaluating}
Bo~Li, Gexiang Fang, Yang Yang, Quansen Wang, Wei Ye, Wen Zhao, and Shikun
  Zhang. 2023{\natexlab{a}}.
\newblock \href {https://doi.org/10.48550/arXiv.2304.11633} {Evaluating
  chatgpt's information extraction capabilities: An assessment of performance,
  explainability, calibration, and faithfulness}.
\newblock \emph{CoRR}, abs/2304.11633.

\bibitem[{Li et~al.(2023{\natexlab{b}})Li, Sun, Tang, Yan, Wu, Huang, and
  Qiu}]{Li2023-codeie}
Peng Li, Tianxiang Sun, Qiong Tang, Hang Yan, Yuanbin Wu, Xuanjing Huang, and
  Xipeng Qiu. 2023{\natexlab{b}}.
\newblock \href {https://doi.org/10.18653/v1/2023.acl-long.855} {{C}ode{IE}:
  Large code generation models are better few-shot information extractors}.
\newblock In \emph{Proceedings of the 61st Annual Meeting of the Association
  for Computational Linguistics (Volume 1: Long Papers)}, pages 15339--15353,
  Toronto, Canada. Association for Computational Linguistics.

\bibitem[{Lou et~al.(2023)Lou, Lu, Dai, Jia, Lin, Han, Sun, and
  Wu}]{Lou2023-usm}
Jie Lou, Yaojie Lu, Dai Dai, Wei Jia, Hongyu Lin, Xianpei Han, Le~Sun, and Hua
  Wu. 2023.
\newblock \href {https://doi.org/10.1609/aaai.v37i11.26563} {Universal
  information extraction as unified semantic matching}.
\newblock \emph{Proceedings of the AAAI Conference on Artificial Intelligence},
  37(11):13318--13326.

\bibitem[{Lu et~al.(2022{\natexlab{a}})Lu, Liu, Dai, Xiao, Lin, Han, Sun, and
  Wu}]{Lu2022-uie}
Yaojie Lu, Qing Liu, Dai Dai, Xinyan Xiao, Hongyu Lin, Xianpei Han, Le~Sun, and
  Hua Wu. 2022{\natexlab{a}}.
\newblock \href {https://doi.org/10.18653/v1/2022.acl-long.395} {Unified
  structure generation for universal information extraction}.
\newblock In \emph{Proceedings of the 60th Annual Meeting of the Association
  for Computational Linguistics (Volume 1: Long Papers)}, pages 5755--5772,
  Dublin, Ireland. Association for Computational Linguistics.

\bibitem[{Lu et~al.(2022{\natexlab{b}})Lu, Liu, Dai, Xiao, Lin, Han, Sun, and
  Wu}]{Lu2022}
Yaojie Lu, Qing Liu, Dai Dai, Xinyan Xiao, Hongyu Lin, Xianpei Han, Le~Sun, and
  Hua Wu. 2022{\natexlab{b}}.
\newblock \href {https://doi.org/10.18653/v1/2022.acl-long.395} {Unified
  structure generation for universal information extraction}.
\newblock In \emph{Proceedings of the 60th Annual Meeting of the Association
  for Computational Linguistics (Volume 1: Long Papers)}, pages 5755--5772,
  Dublin, Ireland. Association for Computational Linguistics.

\bibitem[{Luan et~al.(2018)Luan, He, Ostendorf, and
  Hajishirzi}]{Luan2018-multi}
Yi~Luan, Luheng He, Mari Ostendorf, and Hannaneh Hajishirzi. 2018.
\newblock \href {https://doi.org/10.18653/v1/D18-1360} {Multi-task
  identification of entities, relations, and coreference for scientific
  knowledge graph construction}.
\newblock In \emph{Proceedings of the 2018 Conference on Empirical Methods in
  Natural Language Processing}, pages 3219--3232, Brussels, Belgium.
  Association for Computational Linguistics.

\bibitem[{Ouyang et~al.(2022)Ouyang, Wu, Jiang, Almeida, Wainwright, Mishkin,
  Zhang, Agarwal, Slama, Ray, Schulman, Hilton, Kelton, Miller, Simens, Askell,
  Welinder, Christiano, Leike, and Lowe}]{ouyang2022-training}
Long Ouyang, Jeff Wu, Xu~Jiang, Diogo Almeida, Carroll~L. Wainwright, Pamela
  Mishkin, Chong Zhang, Sandhini Agarwal, Katarina Slama, Alex Ray, John
  Schulman, Jacob Hilton, Fraser Kelton, Luke Miller, Maddie Simens, Amanda
  Askell, Peter Welinder, Paul Christiano, Jan Leike, and Ryan Lowe. 2022.
\newblock \href {http://arxiv.org/abs/2203.02155} {Training language models to
  follow instructions with human feedback}.

\bibitem[{Riedel et~al.(2010)Riedel, Yao, and McCallum}]{Riedel2010-NYT}
Sebastian Riedel, Limin Yao, and Andrew McCallum. 2010.
\newblock Modeling relations and their mentions without labeled text.
\newblock In \emph{Machine Learning and Knowledge Discovery in Databases},
  pages 148--163, Berlin, Heidelberg. Springer Berlin Heidelberg.

\bibitem[{Roth and Yih(2004)}]{Roth2004-conll04}
Dan Roth and Wen-tau Yih. 2004.
\newblock \href {https://aclanthology.org/W04-2401} {A linear programming
  formulation for global inference in natural language tasks}.
\newblock In \emph{Proceedings of the Eighth Conference on Computational
  Natural Language Learning ({C}o{NLL}-2004) at {HLT}-{NAACL} 2004}, pages
  1--8, Boston, Massachusetts, USA. Association for Computational Linguistics.

\bibitem[{Schweter and Akbik(2020)}]{Schweter2020-FLERT}
Stefan Schweter and Alan Akbik. 2020.
\newblock \href {http://arxiv.org/abs/2011.06993} {{FLERT:} document-level
  features for named entity recognition}.
\newblock \emph{CoRR}, abs/2011.06993.

\bibitem[{Song et~al.(2020)Song, Tan, Qin, Lu, and Liu}]{Song2020-MPNet}
Kaitao Song, Xu~Tan, Tao Qin, Jianfeng Lu, and Tie-Yan Liu. 2020.
\newblock \href
  {https://proceedings.neurips.cc/paper_files/paper/2020/file/c3a690be93aa602ee2dc0ccab5b7b67e-Paper.pdf}
  {Mpnet: Masked and permuted pre-training for language understanding}.
\newblock In \emph{Advances in Neural Information Processing Systems},
  volume~33, pages 16857--16867. Curran Associates, Inc.

\bibitem[{Taneeya W.~Satyapanich and Finin(2020)}]{Satyapanich2020-CASIE}
Francis~Ferraro Taneeya W.~Satyapanich and Tim Finin. 2020.
\newblock {CASIE: Extracting Cybersecurity Event Information from Text}.
\newblock In \emph{Proceeding of the 34th AAAI Conference on Artificial
  Intelligence}. AAAI Press.

\bibitem[{Tjong Kim~Sang and De~Meulder(2003)}]{Sang2003-conll03}
Erik~F. Tjong Kim~Sang and Fien De~Meulder. 2003.
\newblock \href {https://aclanthology.org/W03-0419} {Introduction to the
  {C}o{NLL}-2003 shared task: Language-independent named entity recognition}.
\newblock In \emph{Proceedings of the Seventh Conference on Natural Language
  Learning at {HLT}-{NAACL} 2003}, pages 142--147.

\bibitem[{Touvron et~al.(2023)Touvron, Martin, Stone, Albert, Almahairi,
  Babaei, Bashlykov, Batra, Bhargava, Bhosale, Bikel, Blecher, Ferrer, Chen,
  Cucurull, Esiobu, Fernandes, Fu, Fu, Fuller, Gao, Goswami, Goyal, Hartshorn,
  Hosseini, Hou, Inan, Kardas, Kerkez, Khabsa, Kloumann, Korenev, Koura,
  Lachaux, Lavril, Lee, Liskovich, Lu, Mao, Martinet, Mihaylov, Mishra,
  Molybog, Nie, Poulton, Reizenstein, Rungta, Saladi, Schelten, Silva, Smith,
  Subramanian, Tan, Tang, Taylor, Williams, Kuan, Xu, Yan, Zarov, Zhang, Fan,
  Kambadur, Narang, Rodriguez, Stojnic, Edunov, and
  Scialom}]{Touvron2023-llama}
Hugo Touvron, Louis Martin, Kevin Stone, Peter Albert, Amjad Almahairi, Yasmine
  Babaei, Nikolay Bashlykov, Soumya Batra, Prajjwal Bhargava, Shruti Bhosale,
  Dan Bikel, Lukas Blecher, Cristian~Canton Ferrer, Moya Chen, Guillem
  Cucurull, David Esiobu, Jude Fernandes, Jeremy Fu, Wenyin Fu, Brian Fuller,
  Cynthia Gao, Vedanuj Goswami, Naman Goyal, Anthony Hartshorn, Saghar
  Hosseini, Rui Hou, Hakan Inan, Marcin Kardas, Viktor Kerkez, Madian Khabsa,
  Isabel Kloumann, Artem Korenev, Punit~Singh Koura, Marie-Anne Lachaux,
  Thibaut Lavril, Jenya Lee, Diana Liskovich, Yinghai Lu, Yuning Mao, Xavier
  Martinet, Todor Mihaylov, Pushkar Mishra, Igor Molybog, Yixin Nie, Andrew
  Poulton, Jeremy Reizenstein, Rashi Rungta, Kalyan Saladi, Alan Schelten, Ruan
  Silva, Eric~Michael Smith, Ranjan Subramanian, Xiaoqing~Ellen Tan, Binh Tang,
  Ross Taylor, Adina Williams, Jian~Xiang Kuan, Puxin Xu, Zheng Yan, Iliyan
  Zarov, Yuchen Zhang, Angela Fan, Melanie Kambadur, Sharan Narang, Aurelien
  Rodriguez, Robert Stojnic, Sergey Edunov, and Thomas Scialom. 2023.
\newblock \href {http://arxiv.org/abs/2307.09288} {Llama 2: Open foundation and
  fine-tuned chat models}.

\bibitem[{Wadhwa et~al.(2023)Wadhwa, Amir, and Wallace}]{Wadhwa2023-revisiting}
Somin Wadhwa, Silvio Amir, and Byron Wallace. 2023.
\newblock \href {https://doi.org/10.18653/v1/2023.acl-long.868} {Revisiting
  relation extraction in the era of large language models}.
\newblock In \emph{Proceedings of the 61st Annual Meeting of the Association
  for Computational Linguistics (Volume 1: Long Papers)}, pages 15566--15589,
  Toronto, Canada. Association for Computational Linguistics.

\bibitem[{Wang et~al.(2023{\natexlab{a}})Wang, Zhou, Zu, Xia, Chen, Zhang,
  Zheng, Ye, Zhang, Gui, Kang, Yang, Li, and Du}]{Wang2023-instructuie}
Xiao Wang, Weikang Zhou, Can Zu, Han Xia, Tianze Chen, Yuansen Zhang, Rui
  Zheng, Junjie Ye, Qi~Zhang, Tao Gui, Jihua Kang, Jingsheng Yang, Siyuan Li,
  and Chunsai Du. 2023{\natexlab{a}}.
\newblock \href {https://doi.org/10.48550/arXiv.2304.08085} {Instructuie:
  Multi-task instruction tuning for unified information extraction}.
\newblock \emph{CoRR}, abs/2304.08085.

\bibitem[{Wang et~al.(2023{\natexlab{b}})Wang, Zhou, Zu, Xia, Chen, Zhang,
  Zheng, Ye, Zhang, Gui, Kang, Yang, Li, and Du}]{Wang2023}
Xiao Wang, Weikang Zhou, Can Zu, Han Xia, Tianze Chen, Yuansen Zhang, Rui
  Zheng, Junjie Ye, Qi~Zhang, Tao Gui, Jihua Kang, Jingsheng Yang, Siyuan Li,
  and Chunsai Du. 2023{\natexlab{b}}.
\newblock \href {https://doi.org/10.48550/arXiv.2304.08085} {Instructuie:
  Multi-task instruction tuning for unified information extraction}.
\newblock \emph{CoRR}, abs/2304.08085.

\bibitem[{Wang et~al.(2023{\natexlab{c}})Wang, Li, and
  Ji}]{Wang2023-code4struct}
Xingyao Wang, Sha Li, and Heng Ji. 2023{\natexlab{c}}.
\newblock \href {https://doi.org/10.18653/v1/2023.acl-long.202}
  {{C}ode4{S}truct: Code generation for few-shot event structure prediction}.
\newblock In \emph{Proceedings of the 61st Annual Meeting of the Association
  for Computational Linguistics (Volume 1: Long Papers)}, pages 3640--3663,
  Toronto, Canada. Association for Computational Linguistics.

\bibitem[{Wang et~al.(2021)Wang, Jiang, Bach, Wang, Huang, Huang, and
  Tu}]{Wang2021-automated}
Xinyu Wang, Yong Jiang, Nguyen Bach, Tao Wang, Zhongqiang Huang, Fei Huang, and
  Kewei Tu. 2021.
\newblock \href {https://doi.org/10.18653/v1/2021.acl-long.206} {Automated
  concatenation of embeddings for structured prediction}.
\newblock In \emph{Proceedings of the 59th Annual Meeting of the Association
  for Computational Linguistics and the 11th International Joint Conference on
  Natural Language Processing (Volume 1: Long Papers)}, pages 2643--2660,
  Online. Association for Computational Linguistics.

\bibitem[{Yan et~al.(2018)Yan, Tang, Duan, Liu, Wang, Jiang, Zhou, and
  Li}]{Yan2018-assertion}
Zhao Yan, Duyu Tang, Nan Duan, Shujie Liu, Wendi Wang, Daxin Jiang, Ming Zhou,
  and Zhoujun Li. 2018.
\newblock \href {https://doi.org/10.1609/aaai.v32i1.12052} {Assertion-based qa
  with question-aware open information extraction}.
\newblock \emph{Proceedings of the AAAI Conference on Artificial Intelligence},
  32(1).

\bibitem[{Ye et~al.(2022)Ye, Lin, Li, and Sun}]{Ye2022-packed}
Deming Ye, Yankai Lin, Peng Li, and Maosong Sun. 2022.
\newblock \href {https://doi.org/10.18653/v1/2022.acl-long.337} {Packed
  levitated marker for entity and relation extraction}.
\newblock In \emph{Proceedings of the 60th Annual Meeting of the Association
  for Computational Linguistics (Volume 1: Long Papers)}, pages 4904--4917,
  Dublin, Ireland. Association for Computational Linguistics.

\end{thebibliography}
\bibliographystyle{acl_natbib}

\appendix


\end{sloppy}
\end{document}